\documentclass[runningheads]{llncs}
\pdfoutput=1
\usepackage{graphicx}
\usepackage{hyperref}
\usepackage{enumerate}   
\usepackage{xcolor}
\begin{document}
\title{Quantile Encoder: Tackling High Cardinality Categorical Features in Regression Problems}
\titlerunning{Quantile Encoder}
%
\author{Carlos Mougan\inst{1}  \thanks{The author has contributed to this work while he was employed at European Central Bank} \and
David Masip\inst{2}\and
Jordi Nin\inst{3}\and
Oriol Pujol\inst{4}}
%
%
\institute{Electronics and Computer Science, University of Southampton, UK \email{C.Mougan-Navarro@southampton.ac.uk} \and Centre Recerca Matematica - Universitat Autonoma de Barcelona, Spain \email{david26694@gmail.com} \and
 Universitat  Ramon  Llull, ESADE, Barcelona, Spain\\ \email{jordi.nin@esade.edu}\and 
Dpt. of Mathematics and Computer Science - Universitat de Barcelona, Spain \email{oriol\_pujol@ub.edu} }

\maketitle              
\begin{abstract}

Regression problems have been widely studied in machine learning literature resulting in a plethora of regression models and performance measures. However, there are few techniques specially dedicated to solve the problem of how to incorporate categorical features to regression problems. Usually, categorical feature encoders are general enough to cover both classification and regression problems. This lack of specificity results in underperforming regression models. In this paper, we provide an in-depth analysis of how to tackle high cardinality categorical features with the quantile. Our proposal outperforms state-of-the-art encoders, including the traditional statistical mean target encoder, when considering the Mean Absolute Error, especially in the presence of long-tailed or skewed distributions. Besides, to deal with possible overfitting when there are categories with small support, our encoder benefits from additive smoothing. Finally, we describe how to expand the encoded values by creating a set of features with different quantiles. This expanded encoder provides a more informative output about the categorical feature in question, further boosting the performance of the regression model.

\end{abstract}
\section{Introduction}
\label{sec:intro}

In the modeling stage of a machine learning (ML) prediction problem, there is the need of feeding the model with meaningful features that describe the problem in a relevant way. That is why the steps of data preparation and feature engineering are crucial in any ML project~\cite{mle,hands-on}. With the recent amount of available data, there is an inevitable increment in the variety of features. For the specific case of categorical variables this increment has two different effects: (a) the quantity of features is larger, and (b) the number of distinct values found in each feature (cardinality) increases \cite{weworkEncoding}. When facing this second scenario, the problem of representing categorical features effectively and efficiently has a relevant effect on the performance of machine learning model. 

Handling categorical features is a known and very common problem in data science and machine learning, given that many algorithms need to be fed with numerical data~\cite{tutz_2011}. There are many well-known methods for approaching this problem~\cite{Pargent:2019}. However, depending on the kind of problem faced, namely classification or regression problems, some of the techniques for encoding categorical data are more suitable than others. This is particularly true when dealing with large-scale data where errors and outliers are more common and may hinder the computation of reliable statistical measures. 

The most well-known encoding for categorical features with low cardinality is One Hot Encoding \cite{one_hot}. This produces orthogonal and equidistant vectors for each category. However, when dealing with high cardinality categorical features, one hot encoding suffers from several shortcomings \cite{catFeatBayesian}: (a) the dimension of the input space increases with the cardinality of the encoded variable, (b) the created features are sparse - in many cases, most of the encoded vectors hardly appear in the data -, and (c) One Hot Encoding does not handle new and unseen categories.

An alternative encoding technique is Label/Ordinal Encoding \cite{ordinal} which uses a single column of integers to represent the different categorical values. These are assumed to have no true order and integers are selected at random. This encoding handles the problem of the high dimensional encoding found in One Hot Encoding but imposes an artificial order of the categories. This makes it harder for the model to extract meaningful information. For example, when using a linear model, this effect prevents the algorithm from assigning a high coefficient to this feature.

Alternatively, Target Encoding (or mean encoding) \cite{high-cardinality-categorical} works as an effective solution to overcome the issue of high cardinality. In target encoding, categorical features are replaced with the mean target value of each respective category. With this technique, the high cardinality problem is handled and categories are ordered allowing for easy extraction of the information and model simplification. The main drawback of Target Encoding appears when categories with few (even only one) samples are replaced by values close to the desired target. This biases the model to over-trust the target encoded feature and makes it prone to overfitting. To overcome this problem several strategies introduce regularization terms in the target estimation \cite{high-cardinality-categorical}. A possibility is to use an estimator with additive smoothing, such as the M-Estimator, to estimate each category mean.

Although the techniques described before work for both regression and classification techniques, most of the literature focuses on the binary classification tasks even though the meaningful statistics of a binary classification are not well suited for other prediction tasks such as regression or multi-class classification. For example, when dealing with supervised learning regression tasks, calculating the target mean can give a misleading representation of the category due to the statistical properties of the mean if the data shows heavy tails. A reasonable change in this scenario would be the use of other summary statistics more suited to the target distribution. However, to the best of our knowledge, there is no previous research done in using other aggregation statistics other than the mean. In this paper, we study the use of the quantile as a better and more flexible summarizing statistic value on the regression tasks when measuring the Mean Absolute Error in high cardinality datasets. We study its effect in front of skewed and long-tailed distributions. We additionally introduce a regularization strategy to avoid over-representation of the encoded feature when the statistic is computed over a small target subset of data. Moreover, a richer extension of the studied encoder, namely target summaries, that consists of a discrete set of the basic quantile encoder with different hyperparameter values is introduced. 

The strategy is evaluated in different regression scenarios including the presence of outliers, long-tailed, and skewed distributions. We summarize the main contributions of this paper as follows:
\begin{enumerate}[(i)]
\item We define the quantile encoder. This encoder improves the performance of a regression model when evaluated using the Mean Absolute Error. It is also remarkable that the Mean Absolute Error can be improved even when using a different target loss, such as least squares loss. 
\item We show that the quantile encoder improves the performance of regression models when the distribution of the target is long-tailed. 
\item Finally, we introduce the idea of summary encoder, an encoder designed for creating richer representations. This is built by leveraging information from different quantiles.  
\end{enumerate}

For the sake of reproducibility and to help with the development, experimentation, and testing of the methods used in this paper, an open-source python package containing all the implementations used for this paper is released. We refer to this package as \emph{sktools} \cite{sktools}.

The rest of this article is organized as follows. Section \ref{sec:problem_definition} presents a formal definition of the proposed encoding. Section \ref{sec:experimentAndResults} shows the benchmarking datasets, experimental settings, results, and their corresponding discussion. Finally, in Section \ref{sec:conclusion} the main conclusions of the paper are summarized and possible future work is presented.

\section{Quantile encoder}\label{sec:problem_definition}

The problem that we aim to tackle is the improvement of the encoding of categorical variables in regression models. Target encoding with the mean is a valid approach, but not necessarily the most suitable. Target encoding can be easily generalized by replacing the mean with any other summarizing statistic. Thus, following a similar strategy to mean encoding, here we generalize the definition of Target encoding studying the use of the quantile as a summarizing encoding metric in the different categories.


Formally, Quantile encoding is defined as follows: Given a dataset $\mathcal{D}=\{(x_i,y_i)\}, i\in 1\dots N$ with $x_i$ a $d$-dimensional feature vector, $y_i$ its corresponding label, we identify the $j$-$th$ feature from sample $x_i$ as $x_i^{(j)}$. For the following discussion we consider feature $j$-$th$ a categorical variable with $K_j$ different values. The quantile encoder replaces that feature as follows

\begin{equation}
\label{eq:quantile}
    \hat{x}_i^{(j)} = q_p(\{y_k\}) ,\quad \forall \; (x_k,y_k)\in \mathcal{D} \big| \; x_k^{(j)} = x_i^{(j)}, 
\end{equation}

where $q_p$ is the quantile at $p$. Equation \ref{eq:quantile} assigns the $p$ quantile of all targets that share the same categorical value for that feature.

A common issue when using target-based encodings such as mean encoding or the quantile encoding is not having enough statistical mass for some of the encoded categories. And, therefore, this creates features that are very close to the target label. Thus, they are prone to over-fitting. A possible solution is to regularize the target encoding feature using additive smoothing, as in \cite{mestimator,wiki:additive-smoothing}.  To do so, we compute the quantile encoding using the following equation:


\begin{equation}
\label{eqn:mestimate}
\tilde{x}^{(j)}_i = \frac{\hat{x}^{(j)}_i \cdot n_i + q_p(\{y\}) \cdot m }{n_i + m}
\end{equation}
where,
\begin{itemize}
    \item $\tilde{x}^{(j)}_i$ is the regularized Quantile Encoder applied to the value corresponding to element $x_i^{(j)}$.
    \item $\hat{x}^{(j)}_i$ is the non-regularized Quantile Encoder; the value corresponding to element $x_i^{(j)}$ as defined in Equation \ref{eq:quantile}. 
    \item $n_i$ is the number of samples sharing the same value as $x_i^{(j)}$.
    \item $q_p(\{y\})$ is the global p-quantile of the target.
    \item $m$ is a regularization parameter, the higher $m$ the more the quantile encoding feature tends to the global quantile. It can be interpreted as the number of samples needed to have the local contribution (quantile of the category) equal to the global contribution (global quantile).
\end{itemize}

The rationale of Equation \ref{eqn:mestimate} is that, if a class has very few samples, $n_i \ll m$ then the quantile encoding will basically be the global quantile, $\tilde{x}^{(j)}_i \approx q_p(\{y\})$. If a class has a large number of samples, $n_i \gg m$ and $\tilde{x}^{(j)}_i \approx \hat{x}_i^{(j)}$ then the class quantile will have more weight than the global quantile. As a result, the Quantile Encoder transformation of categorical features has two different hyperparameters that can be tuned to increase, adjust, and modify the type of encoding. These are the following:


\begin{itemize}
    \item $m$ is a regularization hyperparameter. The range of this parameter is $m \in [0, \infty)$. For the specific case where $m=0$, there is no regularization. The larger the value of $m$, the most the Quantile Encoder features tends to the global quantile. 
    
    \item $p$ is the value of the quantile of the target probability distribution. The range of this parameter is $p \in [0, 1]$. When $p$ is 0.5, we obtain the median encoder, as it applies $q_{0.5}$, the median of the target in each category.
\end{itemize}

\subsection{Summary encoder}

A generalization of the quantile encoder is to compute several features corresponding to different quantiles per each categorical feature, instead of a single feature. This allows the model to have broader information about the target distribution for each value of that feature than just a single value. This richer representation will be referred to as {\it summary encoder}. Formally, it is defined as

\begin{equation}
    \hat{x}_i^{(j)} = \{q_{p_m}(\{y_k\})\} ,\quad \forall \; (x_k,y_k)\in \mathcal{D} \big| \; x_k = x_i^{(j)}, \; m=1\dots M 
\end{equation}

where $M$ is the number of new features, and $p_m$ are the values of the quantiles to use. This representation changes a single feature by a set of $M$ quantiles according to the values in $p_m$.

\section{Experiments}\label{sec:experimentAndResults}

To perform an empirical evaluation of the proposed statistical encoding techniques, we developed a framework to ensure that all experiments described in this paper are fully reproducible. Original data, data preparation routines, code repositories, and methods are publicly available at \cite{qe_experiments}. Experiments have been organized into three groups: Firstly, we assess the performance of quantile encoder when compared with the state-of-the-art encodings, namely catboost, M-estimate, target, James-Steiner encoder, Generalized Linear Mixed Model Encoder, and ordinal encoder. To do this comparison we used the Mean Absolute Error. In the second group of experiments, we aim at showing the dependence of the encoding on the evaluation metric. To that effect, we study the performance of the quantile encoder when used with a least-squares loss model in terms of mean absolute error and mean squared error. Finally, we compare summary encoder with quantile and target encoders, the goal of these last experiments is to create more informative encoders to be able to boost regression algorithms performance.

\subsection{Dataset bench-marking}\label{sec:datasets}

The scenario we are addressing is characterized by datasets that display categorical features with high cardinality and skewed/long tail target distribution in a regression environment. Unfortunately, most of the machine learning benchmarking datasets do not display these common features of many real-life problems. Thus, following the open data for reproducible research guidelines described in \cite{turing} and for measuring the performance of the proposed methods, we have used a synthetic dataset for a more theoretical evaluation together with 4 open-source datasets for an empirical comparison. The selected datasets are:

\begin{itemize}
    \item \textit{The StackOverflow 2018 Developer survey}\label{dataset:StackOverflow2018} \cite{so2018} is a data set with only five categorical features, namely country, employment status, formal education, developer type, and languages worked. The target variable is the annual salary of the user. 
    
    \item \textit{The StackOverflow 2019 Developer survey}\label{dataset:so2019} \cite{so2019} is a data set with a single categorical feature (country), few numerical features (working week hours, years of coding, and age), and a long-tailed target variable (salary).
    
    \item \textit{Kickstarter Projects} \cite{ksdata} is a data set with crowd-funding projects where the goal is to predict what is the funding goal of each project. The categorical features are the crowd-funding type, the country, the state, and the currency.
    
    \item \textit{Medical Payments}\cite{medical_payments} \label{dataset:medical} is a data set with information about the price of a medical treatment. The dataset consists of 10 categorical features containing information about the state, city, zip code, country, physician type, physician country, the payment method, and its nature.
\end{itemize}

The datasets have undergone a minimal curation process, where miscellaneous features are removed and only the columns that are considered meaningful and informative for the modeling of the problem are kept. 

For the synthetic dataset, we have used the Cauchy distribution in Eq.~\ref{eqn:cauchy} to create a target distribution with long tails. The Cauchy distribution is parameterized by $t$ and $s$, being $t$ the location parameter and $s$ the scale parameter as follows,

\begin{equation}
\label{eqn:cauchy}
    Cauchy(x;t,s) = \frac{1} {s\pi(1 + ((x - t)/s)^{2})}.
\end{equation}

Next,we have created two features, $x_1$ and $x_2$ accordingly,

$$
c_i \sim U(0, 100), \quad x_1 \sim Cauchy(x;c_i,1), \quad x_2 \sim Cauchy(x;c_i,2),
$$

where $\epsilon \sim N(0, 1)$ is sampled from a normal distribution and $x_1$ and $x_2$ are generated sampling from the Cauchy density function. Both features even if numerical are then treated as categorical variables. The corresponding regression target is generated as

$$
y = x_1 + x_2 + \epsilon
$$

\subsection{Code and reproducibility}

For the sake of reproducibility, the code for the experiments has been encapsulated in a library, \emph{sktools} \cite{sktools}. It can be found \href{https://sktools.readthedocs.io/}{https://sktools.readthedocs.io/}. The library contains the implementations presented in this paper such as Quantile Encoder regularized with an additive smoothing and the Summary Encoder.  Notebook and experiments are hosted on the following \href{https://github.com/david26694/QE\_experiments}{Github repository} \cite{qe_experiments}. 

With respect to quantile encoder the default values for $m$ and $p$ are $m=1$ and  $p=0.5$. Default values for the hyperparameters of the summarizing encoder are $m=100$ and defines three encoding features containing the quantiles at $0.4, 0.5$ and $0.6$.

As a final estimator we have used a Generalized Linear Model with an reguralarization hyperparameter ($l1\_penalty$) that we have optimized across the crossvalidation folds.

\subsection{Comparison of all encoding methods}

This first experiment consists of comparing the performance of the quantile encoder against state-of-the-art encoding techniques in terms of the Mean Absolute Error (MAE). To do so, we have used the methods implemented in the Category Encoders library \cite{category_encoders} to evaluate them against our proposed encoding technique. This set of encoding techniques includes ordinal encoding, James-Stein Encoder \cite{jamesstein_1,jamesstein_2,jamesstein_3}, and several state of the art target encodings with different regularization values such as catboost \cite{catboost-encoder}, classical target encoder \cite{high-cardinality-categorical,category_encoders} and M-estimate target encoding \cite{mestimator} and generalized linear mixed models encoder \cite{mixed_models,category_encoders}. Every experiment has been executed using 3 times repeated 4-fold cross-validation on the parameters of each method.

\begin{figure}[ht]
    \centering
    \includegraphics[width=0.9\textwidth]{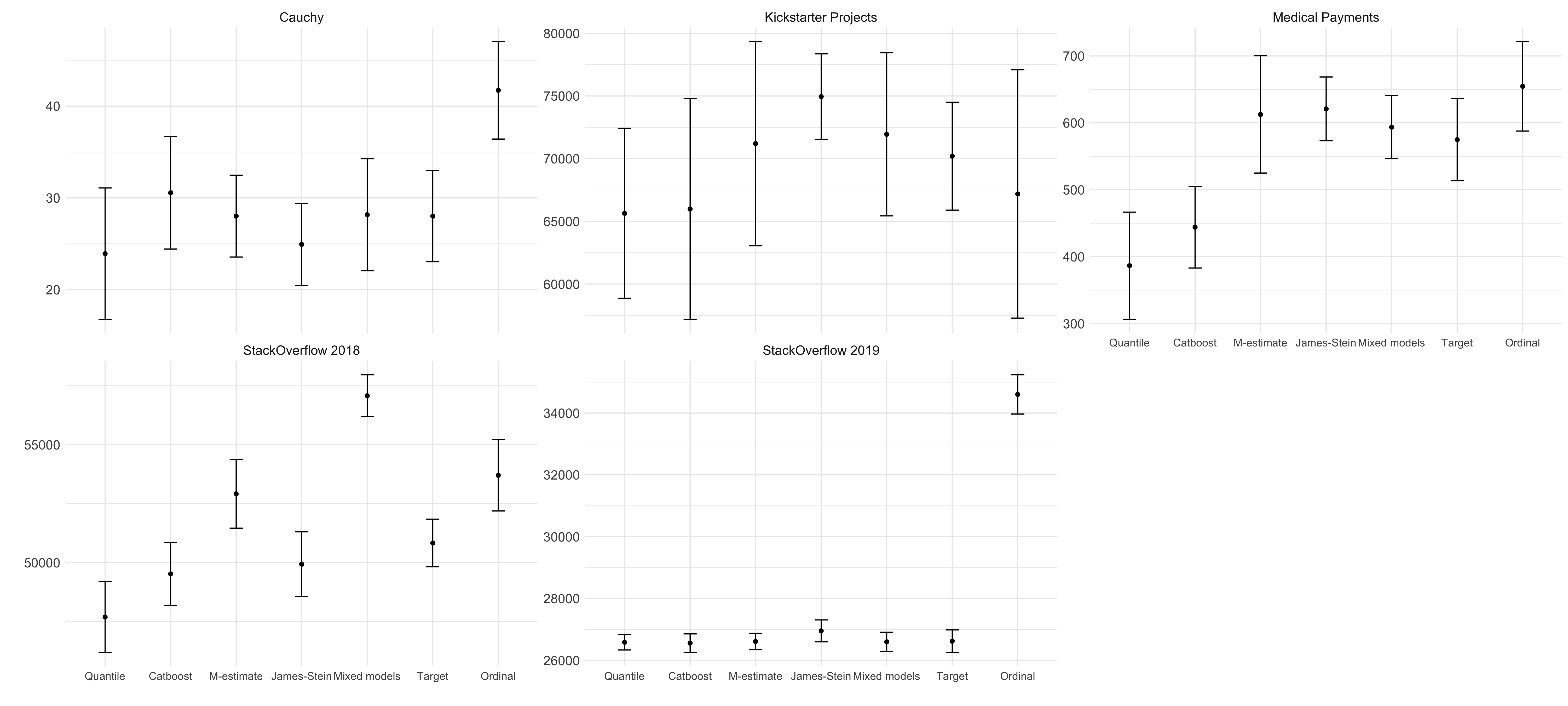}
    \caption{Comparison between Quantile Encoder and several categorical encoding techniques using the cross validated MAE error.}
    \label{fig:categories}
\end{figure}

Figure \ref{fig:categories} shows the results of the comparison on the cross-validated sets for the aforementioned datasets. Observe that on average, the quantile encoder achieves the best scores, followed by catboost. In all datasets, Quantile Encoder performs in the worst-case scenario similar to other state-of-the-art techniques. However, in three out of five cases it produces a solid improvement of the Mean Absolute Error. As expected, ordinal encoder yields the worst performance in these experiments. One of the advantages of using the quantile to encode categorical features over the mean encoding is that it allows us to have one more tunable hyperparameter.  

\subsection{Encoding dependence with respect to the evaluation metric}

When evaluating machine learning regression models, the following natural question arises \textit{For which metrics does the encoding technique give an improvement with respect to the alternatives?}

The mean is the estimator that minimizes the Mean Squared Error (MSE), meanwhile for the Mean Absolute Error is the median~\cite{Jaynes2003,median_proof_ppt}. This statement supports the hypothesis that the median encoder may improve the performance of any regression model when it is measured with the MAE. To provide empirical evidence we evaluate mean and quantile encoders in front of MAE and MSE evaluation metrics. It is important to highlight that MAE error has one advantage versus the MSE from an interpretation point of view, in the sense that MAE maintains the units of the quantities giving a more intuitive representation of the performance of the model to the user. We evaluate both encodings with both metrics using an Elastic Net~\cite{eNet,eNet2} with default scikit learn hyperparameters ($alpha=1.0$, $l1\_ratio=0.5$) as estimator. Hyperparameters are optimized using a grid search with parameters $m \in \{0,1,10,50 \}$ and $quantile \in \{0.25,0.5,0.75\}$. 



\begin{table}[]
\begin{tabular}{cc}
\begin{minipage}[t]{.6\linewidth}
\vspace{0pt}
\centering
\includegraphics[width=0.9\textwidth]{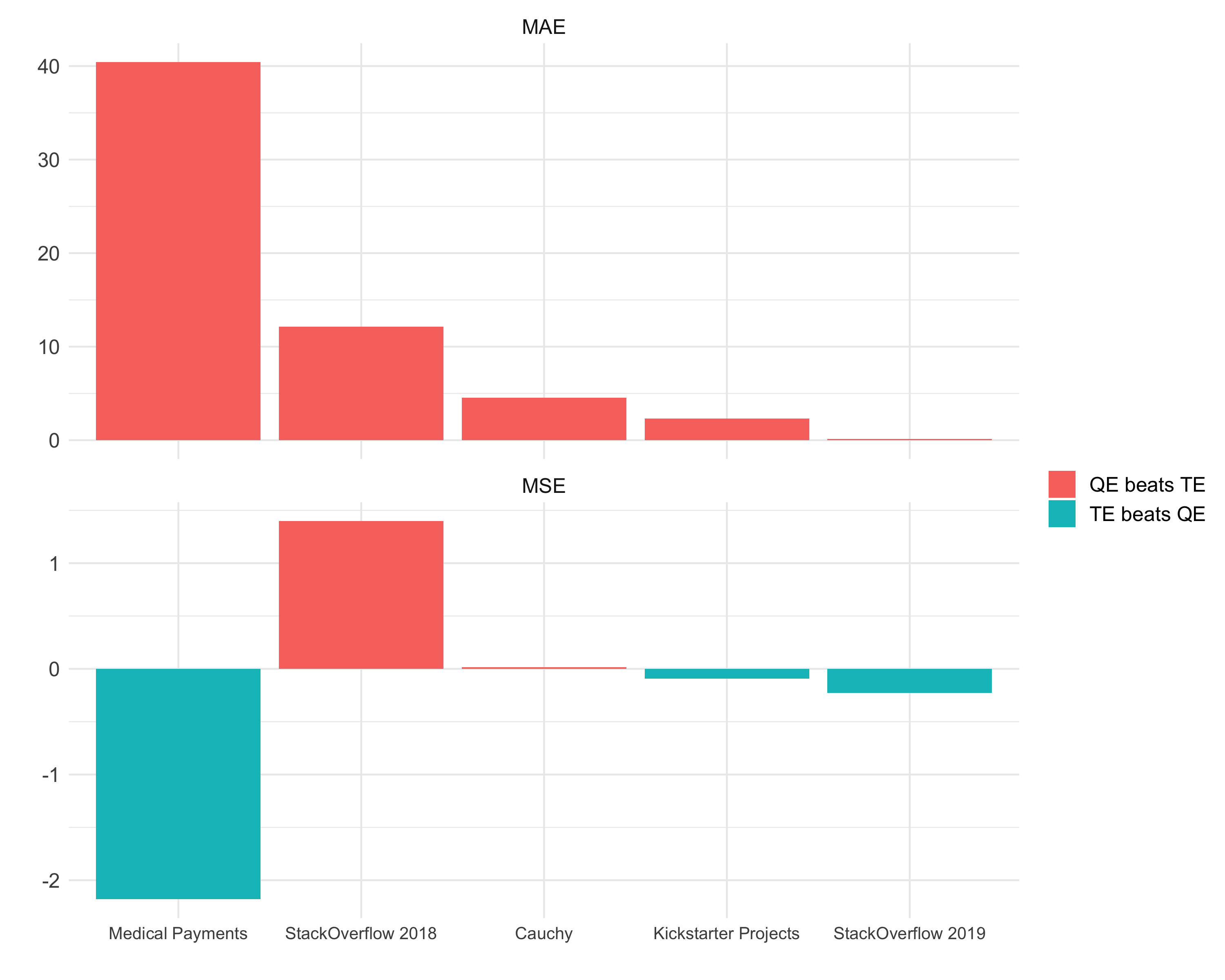}
\label{fig:MAEvsMSE}
\end{minipage}%
&
\begin{minipage}[t]{.3\linewidth}
\vspace{50pt}
\centering
\begin{tabular}{lrr}
  \hline
Dataset & p-value & $P_Q$\\ 
  \hline
Cauchy & 0.0881 & 0.667\\ 
  Kickstarter Projects & 0.0461 & 0.667\\ 
  Medical Payments & 0.0002 & 1.000\\ 
  StackOverflow 2019 & 0.3955 & 0.583\\ 
  StackOverflow 2018 & 0.0002 & 1.000\\ 
   \hline \label{tab:Wilcoxon}
\end{tabular} 

\end{minipage}\\

(a)&(b)
\end{tabular}
    
    \caption{(a) Comparison between Quantile Encoder and Target Encoder for different evaluation metrics, (b) Wilcoxon's test p-values.}
    \label{fig:Wilcoxon}
\end{table}

Figure in Table \ref{fig:MAEvsMSE}(a) shows the percentual difference between each encoding with respect to two metrics. In the upper part of the figure, we can see that the Quantile Encoder achieves better results than the Target Encoder for all datasets except for one when measuring the MAE metric. The encoding yields bigger percentual differences in medical payments and the StackOverflow datasets. In the lower part of the figure, we have the same plot using the MSE metric. Observe that mean encoder achieves better results in three out of five experiments. Observe that in this last case, percentual performance differences are smaller than in the case of MAE. Additionally, quantile encoder performs better than mean encoder in two of them. It is worth noting that the loss function of elastic net corresponds to a least-squares loss. Thus, this should benefit mean encoding and harm the performance of quantile encoder. However, results show the robustness of the quantile encoder even when in this adversarial case.

To verify the generalization of this observation, a quantile encoder is statistically validated on the selected data sets. The null hypothesis states that quantile encoder and mean encoder has the same performance when considering MAE. The p-value in Table \ref{fig:Wilcoxon}(b) shows the results of the Wilcoxon test \cite{Wilcoxon1992}~\footnote{The Wilcoxon test is a non-parametric statistical hypothesis test used to compare two repeated measurements on a single sample to assess whether their population means ranks differ.} on the MAE on 3 repetitions of 4-fold cross-validation. Observing Table~\ref{fig:MAEvsMSE}(b) we see that the p-values in 3 out of 5 datasets are able to reject the null hypothesis at a significance level of $0.05$. We observe that in the \textit{cauchy} dataset the rejection level is found at a significance level of 0.10.  Finally, in the 2019 StackOverflow dataset, we are not able to reject the null hypothesis. Despite this last result, the quantile encoder is not worse in any of the five datasets. This shows that the quantile encoder is a useful technique to encode categorical variables when optimizing the MAE. We additionally compute the probability of the quantile encoder outperforming the target encoder. This is shown in the column $P_Q$ of Table~\ref{tab:Wilcoxon}. The value is computed by computationally estimating the empirical distribution of the difference of the performance values using kernel density estimation and integrating the area of the distribution where quantile encoder outperforms target encoding. The obtained results are in accordance with the Wilcoxon test. Note that in all datasets there exists a larger probability that the quantile encoder outperforms target encoding.



\subsection{Summary encoder performance}

The summary encoder method provides a broader description of a categorical variable than the quantile encoder. In this experiment, we empirically verify the performance of both in terms of their MAE when they are applied to different datasets. For this experiment we have chosen 3 quantiles that split our data in equal proportions for the summary encoder, {\em i.e.}, $p=0.25$, $p=0.5$ and $p=0.75$.

\begin{figure}[ht]
    \centering
    \includegraphics[width=0.85\textwidth]{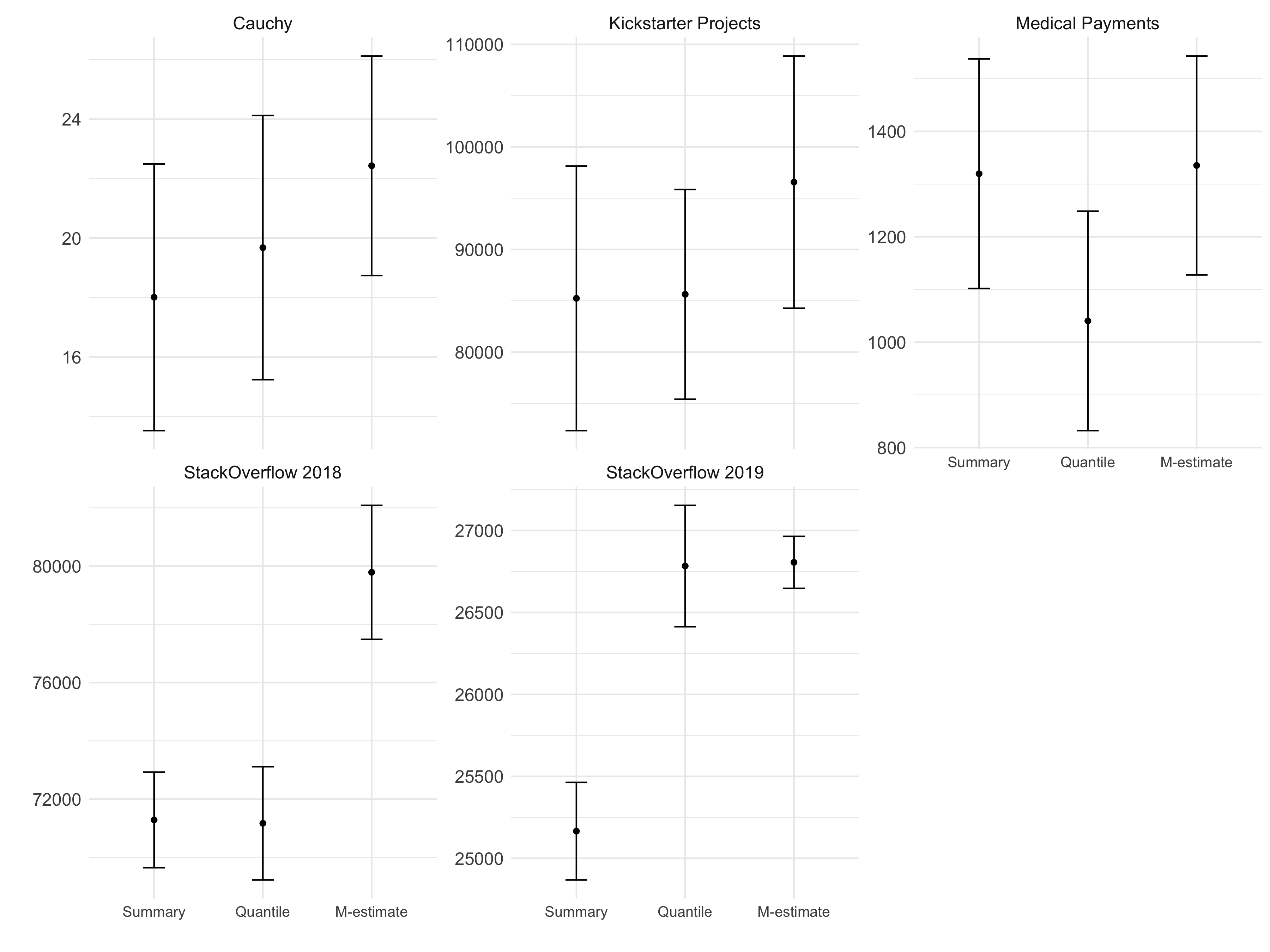}
    \caption{Comparison between Summary, Quantile and Target encoders using the cross validated MAE error.}
    \label{fig:summary}
\end{figure}

Figure \ref{fig:summary} depicts the results for this experiment. Notice that the mean performance of the summary encoder suggests a better performance when compared to the target encoder. The same behavior is observed when compared with quantile encoder in some cases. It must be noted that some extra caution needs to be taken when using the summary encoder as there is more risk of overfitting the more quantiles are used. This usage of the Quantile Encoding requires more hyperparameters as each new encoding requires two new hyperparameters, $m$ and $p$, making the hyperparameter search computationally more expensive.

\subsection{Discussion}

The experiments show that quantile encoder represents better high cardinality categorical data in several scenarios. The observed improvements are:

\begin{itemize}
    \item Quantile encoder is robust in front of outliers. On the contrary, target encoding is very sensitive to samples in the training set with extreme values.
    \item From an optimization point of view, the mean is the estimator that minimizes the MSE of a sample. On the contrary, and besides optimizing MAE, the use of quantile encoder is a sensible option for general use as it provides a highly tunable summary statistic suited to a broader set of metrics. Besides, from a regression point of view, MAE is a more intuitive metric that helps users interpreting the results.
    \item Finally, quantiles can be grouped to provide a much richer description of a categorical feature. For instance, we can run the percentiles 25, 50, and 75, which give much more information than just computing the mean. More features provide more information to the model. However, more features also increase the risk of overfitting and the problem starts gaining dimensionality. With the use of the Summary Encoder dimensionality does not become a hazard such as in the case of one-hot encoder. Nonetheless, the regularization techniques are to be considered to avoid overfitting in this case. 
\end{itemize}

\section{Conclusion} \label{sec:conclusion}

In this article, we have studied the quantile encoder. We make three contributions related to the encoding of categorical features with high cardinality in regression models. Our first contribution is the definition of the Quantile Encoder as a way to encode categorical features in noisy datasets in a more robust way than mean target encoding. Quantile Encoding maps categories with a more suitable statistical aggregation than the rest of the compared encodings when categories display in long-tailed or skewed distributions. To provide empirical evidence we benchmark the approach in different datasets and provide statistics that support our claims. The second contribution is the observation that categorical encodings are sensitive to the model's loss function and interpretation/evaluation performance metric. In this respect, the performance of the model can heavily change if a general or not correctly selected encoder is chosen. In our case, quantile encoder is suitable when using mean absolute error as an evaluation metric. Finally, due to the tunable hyperparameters of the quantile encoder, this shows a large versatility, being used for different metrics. Additionally, the concatenation of different quantiles allows for a wider and richer representation of the target category that results in a performance boost in regression models. To aid in the goal of open-source and reproducible research, we have released a toolkit, namely \emph{sktools} \cite{sktools}, as an open-source Python package that provides a flexible implementation of the concepts introduced in this paper. For the summary encoder, we have used the M-estimate regularization technique, but further research can be done in the path of avoiding overfitting when creating a set of features out of a high-cardinality categorical feature. Strategies such as those found in leave-one-out encoding, or catboost encoder \cite{catboost-encoder} could be considered to that effect. 


%
%
%
%
%

\subsection*{Acknowledgements}
This  work  was  partially  funded  by  the  European  Commission  under  contract numbers NoBIAS — H2020-MSCA-ITN-2019 project GA No. 860630. \\
This work has been partially funded by the Spanish project PID2019-105093GB-I00 (MINECO/FEDER, UE)
\bibliographystyle{spmpsci}      
\bibliography{sample}   

\end{document}